\documentclass{llncs}

\usepackage{amsmath}
\usepackage{amssymb}
\usepackage{caption}
\usepackage{float}
\usepackage{graphicx}
\usepackage{hyperref}
\usepackage{multirow}
\usepackage{subfig}

\usepackage{amsmath}
\usepackage{bbm}
\usepackage{amssymb}

\begin{document}
\title {Deeper Image Quality Transfer: Training Low-Memory Neural Networks for 3D Images}
\author{Stefano B. Blumberg\inst{2}, Ryutaro Tanno\inst{2},  \\Iasonas Kokkinos\inst{2}, Daniel C. Alexander\inst{1,2}}

\institute{Clinical Imaging Research Centre, National University of Singapore
\and Department of Computer Science and Centre for Medical Image Computing, \\ University College London (UCL)}
\maketitle    

\begin{abstract}
In this paper we address the memory demands that come with the processing 
of 3-dimensional, high-resolution, multi-channeled medical images in deep learning.
We exploit memory-efficient backpropagation techniques, to reduce the memory complexity of network training from being linear in the network's depth, to being roughly constant -- permitting us to elongate deep architectures with negligible memory increase. 
We evaluate our methodology in the paradigm of Image Quality Transfer, whilst noting its potential application to various tasks that use deep learning.  We study the impact of depth on accuracy and show that deeper models have more predictive power, which may exploit larger training sets. We obtain substantially better results than the previous state-of-the-art model with a slight memory increase, reducing the root-mean-squared-error by 13\%. Our code is publicly available.
\end{abstract}

\section{Introduction}
Medical imaging tasks require processing high-resolution (HR), multi-channeled, volumetric data, which produces a large memory footprint. 
Current graphics processing unit (GPU) hardware limitations, constrain the range of models that can be used for medical imaging, since only moderately deep 3D networks can fit on common GPU cards during training. 
Even with moderately deep networks, current practice in medical imaging involves several compromises, such as utilising a small input volume e.g. patches \cite{BIQTCNN}, that forces the network to perform local predictions, or by using a small minibatch size \cite{VNET}, which can destabilise training.  Whilst the impact of network depth has been extensively demonstrated to produce improved results in computer vision \cite{LSMN,DRLIR}, this issue has attracted scant attention in medical image computing, due to the aforementioned limitations. 

We introduce memory-efficient backpropagation techniques into medical imaging, where elongating a network produces a negligible memory increase, thus facilitating the training of deeper and more accurate networks.  We combine two memory-efficient learning techniques: checkpointing \cite{TDNSMC} and reversible networks (RevNets) \cite{RAADRNN,RRN}, that exchange training speed with memory usage.  Deepening an existing architecture, we systematically demonstrate that elongating a network increases its capacity, unleashing the potential of deep learning. 
\\ \indent We demonstrate the effectiveness of this technique within the context of Image Quality Transfer (IQT) \cite{IQTADMRI}. IQT is a paradigm for propagating information from rare or expensive high quality images (e.g. from a unique high-powered MRI scanner) to lower quality but more readily available images (e.g. from a standard hospital scanner).  We consider the application of IQT to enhance the resolution of diffusion magnetic resonance imaging (dMRI) scans -- which has substantial downstream benefits to brain connectivity mapping \cite{BIQTCNN,IQTADMRI}.
\\ \indent By studying the impact of network depth on accuracy, we demonstrate that deeper models have substantially more modelling power and by employing larger training sets,  we demonstrate that increased model capacity produces significant improvements (Fig.\ref{Table_combined}).  We surpass the previous state-of-the-art model of \cite{BIQTCNN}, reducing the root-mean-squared-error (RMSE) by 13\% (Fig.\ref{Table_combined}) -- with negligible memory increase (Fig.\ref{Table_MemEff}). 
\\ \indent We expect that our methods will transfer to other medical imaging tasks that involve volume processing or large inputs, e.g. image segmentation \cite{EMS3D}, synthesis \cite{wolterink2017} and registration \cite{FPIR} -- therefore our implementation is publicly available at \url{http://mig.cs.ucl.ac.uk/}. 

\newcommand{\refsec}[1]{Sec.~\ref{#1}}


\section{Memory-Efficient Deep Learning }
In this section, we use the concept of a computational graph to explain how the memory consumption of backpropagation increases with deeper networks.  We present RevNets and illustrate how to insert them in a pre-defined architecture.  Finally we combine checkpointing with this elongated system, to perform manual forward and backward propagation, allowing us to trade memory consumption with computational cost during training. 


\noindent \\ \textbf{Memory Usage in Backpropagation} We consider a chain-structured network with sets of neurons organized in consecutive layers $X^{1},X^{2},\ldots,X^{N}$, related in terms of  non-linear functions, $X^{i+1} = f^{i}(X^{i},\theta^{i})$, with parameters $\theta^{i}$ specific to layer $ i $. Training aims at minimizing a loss function, $L(X^{N},Y)$, where $Y$ is the target output -- in the setting of IQT, the high-quality patch. 
\\ \indent  Backpropagation  recursively computes the gradient of the loss with respect to the parameters $\theta^{i}$ and neuronal activations $ X^{i}_{0} $, at layer $i $.
Its computation at layer $i$ takes inputs $ \frac{d f^{i}}{dX^{i}}, \frac{d L}{dX^{i+1}},  X^{i}_{0} $ to compute $ \frac{d L}{d \theta^i}|_{X^{i}_{0}}, \frac{d L}{d X^i} |_{X^{i}_{0}}$. Therefore backpropagating from $X^N$ to $X^1$ requires all intermediate layer activations $X^{1}_{0},\ldots, X^{N}_{0} $. This means that memory complexity can scale linearly in the network's depth, which is the case in standard implementations.
\\ \indent Memory-efficient variants of backpropagation trade computational time for training speed without sacrificing accuracy. As an example, when backpropagating at layer $i$, one can compute $X^{i}_{0}$ from scratch, by re-running a forward pass from the input $X^{0}_{0}$ to $X^{i}_{0}$. The memory usage is constant in network depth, but the computational cost is now quadratic in depth. Checkpointing \cite{TDNSMC} -- a method that applies to general graphs, allows the memory cost to scale at square root of the network's depth, whilst increasing the computational cost by a factor of $ \frac{3}{2} $.  RevNets \cite{RAADRNN,RRN}, also increase the computational cost by a similar factor, via their invertibility, we may keep the memory consumption constant in the network's depth. In our implementation we use a combination of the two methods, as detailed below.

\begin{figure}[ht]
\centering
\includegraphics[width=1.0\linewidth]{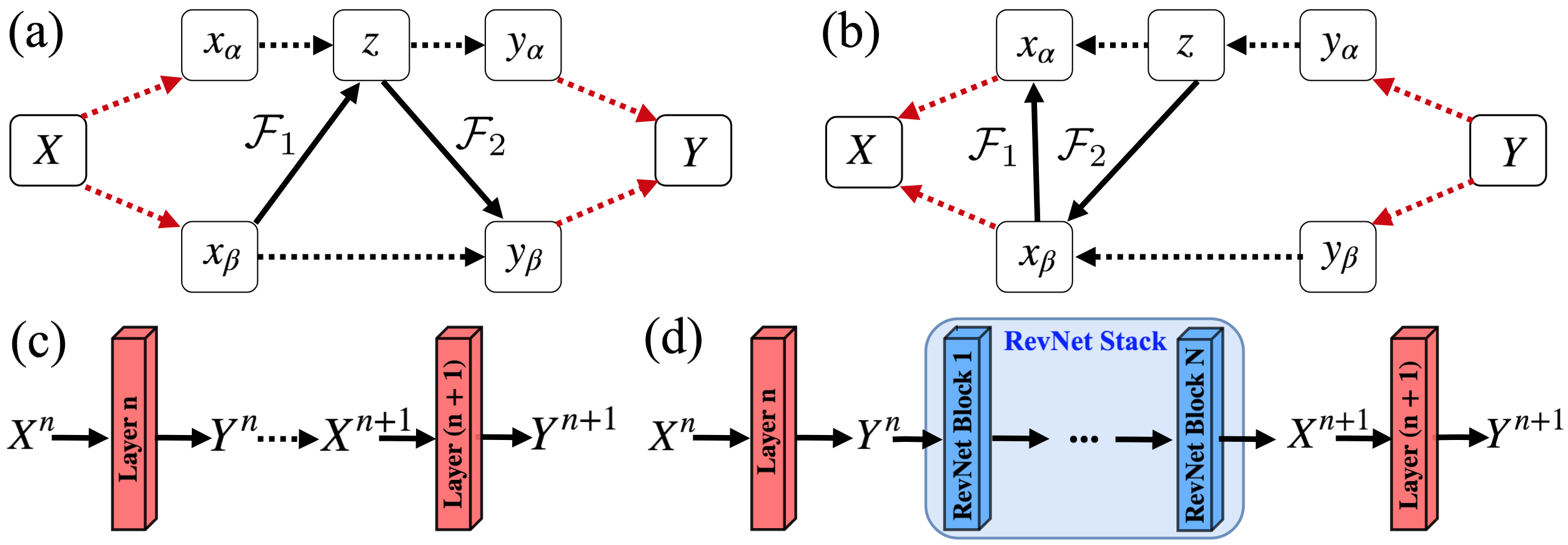}
\caption{A RevNet (top) and  architecture elongation (bottom). Top: Flowcharts of a RevNet block \cite{RAADRNN,RRN}, with input and output feature maps respectively $ X,Y $ in (a) forward pass and (b) backward pass.  Black dotted lines are identity operations, red dotted lines are concatenation and splitting operations, $\mathcal{F}_{1},\mathcal{F}_{2} $ are non-linear functions. Bottom: We elongate a network by inserting $ N $ RevNet blocks between layers $ n,n+1 $ of a neural network. First, as in (c) we split the intermediate activation between layers  $ n,n+1 $ into two computational nodes $ Y^{n},X^{n+1} $; then, as in (d), we insert $ N $ RevNet blocks between $ Y^{n},X^{n+1} $.}\label{RN_EXT}
\end{figure}

\noindent \\ \textbf{Reversible Networks} A RevNet \cite{RAADRNN,RRN} is a neural network block containing convolutional layers, where its input activations can be computed from its output activations (Fig.\ref{RN_EXT}b).  We use two residual function bottlenecks $ \mathcal{F}_{1},\mathcal{F}_{2} $ \cite{DRLIR} as its convolutional blocks.  When stacking RevNet blocks, we only cache final activations of the entire stack.  During backpropagation, we compute intermediate stack activations on-the-fly, via the inversion property.  The formulae for the forward and backward (inversion) are:

\begin{align*}
   \textbf{For}&\textbf{ward Pass}                 & \textbf{Inv}&\textbf{ersion}
\\  X &= [x_{\alpha}, x_{\beta}]                   & Y &= [y_{\alpha}, y_{\beta}] 
\\  z &= x_{\alpha} + \mathcal{F}_{1}(x_{\beta})   & z &= y_{\alpha}
\\  y_{\beta} &= x_{\beta} + \mathcal{F}_{2}(z)    & x_{\beta} &= y_{\beta} - \mathcal{F}_{2}(z)
\\  y_{\alpha} &= z								   & x_{\alpha} &= z - \mathcal{F}_{1}(x_{\beta})
\\  Y &= [y_{\alpha}, y_{\beta}]				       & X &= [x_{\alpha}, x_{\beta}]
\end{align*} 
\noindent \textbf{Augmenting Deep Neural Networks}  Suppose we wish to improve the performance of a neural network architecture by making it deeper \cite{LSMN}.  We propose to pick two layers of the architecture and add a stack of RevNets between them (Fig.\ref{RN_EXT}c,d). This refines the intra-layer connection and facilitates a more complicated mapping to be learnt between them. 



\begin{figure}[h]
\begin{center}
\includegraphics[width=\linewidth, height=60pt]{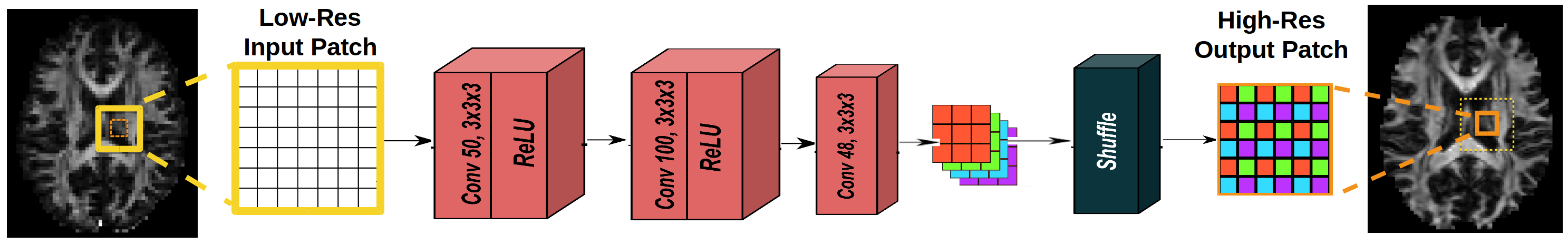}\caption{2D illustration of the baseline network: 3D ESPCN \cite{BIQTCNN}.}\label{ESPCN}
\end{center}
\end{figure}
\begin{figure}[h]
\begin{center}
\includegraphics[width=\linewidth , height=175pt]{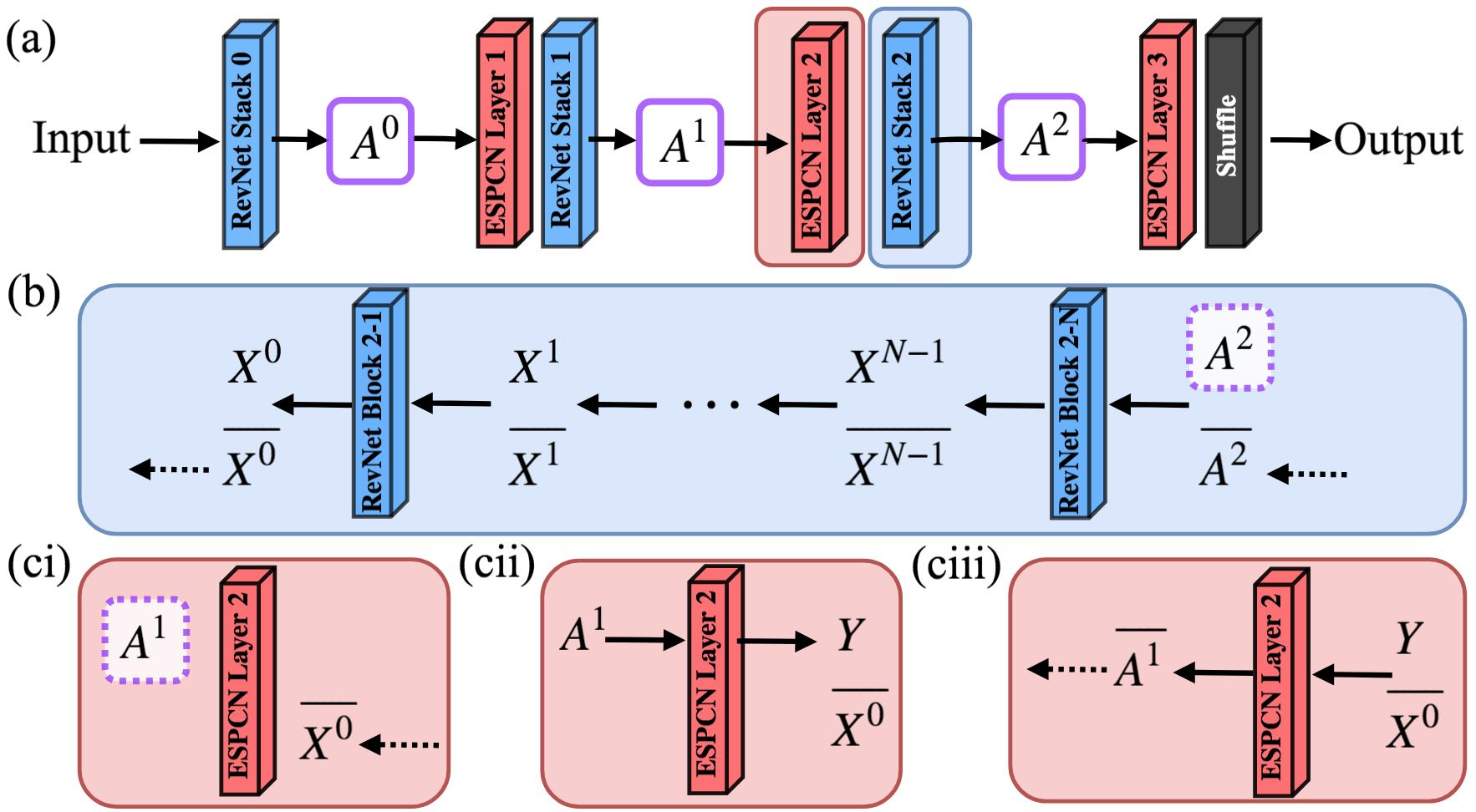}
\caption{ We augment the ESPCN (Fig.\ref{ESPCN}) and illustrate (a) the global forward pass, (b,c) the backward pass on part of the network. (a) Augment the ESPCN (\ref{RN_EXT}c,d). In the forward pass, we cache (purple squares) activations $ A^{0}, A^{1}, A^{2} $ and create no computational graphs. (b,c) We illustrate backpropagation in the section between the activations $ A^{1}, A^{2} $: (b) Load $ A^{2} $ from the cache (purple dotted square) and receive the loss gradient $ \overline{A^{2}} := \frac{\partial L}{\partial A^{2}} $ from ESPCN Layer 3 (dotted arrow).  Iteratively we pass the activation and gradient backwards per block, deleting redundant values.  The final gradient $ \overline{X^{0}} $ is passed to ESPCN Layer 2 (dotted arrow).  (c) Backpropagation on ESPCN Layer 2.  (ci) Load activation $ A^{1} $ from the cache (purple dotted square) and $ \overline{X^{0}} $ is passed from RevNet Block 2-1 (dotted arrow) (cii) Create a computational graph through ESPCN Layer 2.  (ciii) Combine the computational graph with $ \overline{X^{0}} $ to backpropagate backwards on the ESPCN Layer 2.  Finally pass the gradient $ A^{1} $ to RevNet Stack 1 (dotted arrow).}\label{ESPCN_RN_FP_BP}
\end{center}
\end{figure}

\noindent \\\textbf{Augmenting the ESPCN} 
We evaluate our procedure with the ESPCN network, which holds the benchmark for HCP data super-resolution (IQT) \cite{BIQTCNN}.  The ESPCN (Fig.\ref{ESPCN}) is a simple four-layer convolutional neural network, followed by a shuffling operation from low-to-high-resolution space: a mapping $H \times W \times D \times r^3 C \rightarrow rH \times rW \times rD \times C$, with spatial and channel dimensions respectively $ H,W,D $ and $ C $.
\\ \indent We augment the ESPCN (Fig.\ref{ESPCN_RN_FP_BP}a), by adding $ N $ RevNet blocks in a stack, preceding each ESPCN layer.  When optimising network weights, we can either perform the forward and backward pass (backpropagation) via the standard implementation i.e. a single computational graph, which we denote as ESPCN-RN-N-Naive; or utilise the reversibility property of RevNets with a variant of checkpointing, denoted by  ESPCN-RN-N and illustrated in Fig.\ref{ESPCN_RN_FP_BP}. Note ESPCN-RN-N-Naive, ESPCN-RN-N have identical predictions and performance (e.g. in Fig.\ref{Table_combined}), only the computational cost and memory usage differ (Fig.\ref{Table_MemEff}).  We finally note that this technique is not restricted to the ESPCN or to super-resolution, but may be employed in other neural networks.

\section{Experiments and Results}
\textbf{IQT} We formulate super-resolution as a patch-regression, to deal with its large volume, where the input low-resolution (LR) volume is split into overlapping smaller sub-volumes and the resolution of each is sequentially enhanced \cite{BIQTCNN,IQTADMRI}. The HR prediction of the entire 3D brain is obtained by stitching together all the corresponding output patches (Fig.\ref{Results_pic}).
\\
\\ \textbf{HCP Data} We follow \cite{BIQTCNN,IQTADMRI} and utilise a set of subjects from the Human Connectome Project (HCP) cohort \cite{ADMRI}.  This involves healthy adults (22-36 years old), where we specifically vary race, gender and handedness, which effects brain structure.  Each subject's scan contains 90 diffusion weighted images (DWIs) of voxel size $ 1.25^{3}\text{mm}^3 $ with $ b = 1000 \text{s/mm}^2 $.  We create the training-and-validation-set (TVS) by sampling HR sub-volumes from the ground truth diffusion tensor images (DTIs, obtained from DWIs) and then down-sampling to generate the LR counterparts. Down-sampling is performed in the raw DWI by a factor of $ r=2 $ in each dimension by taking a block-wise mean, where $ r=2 $ is the up-sampling rate and then the diffusion tensors are subsequently computed. Lastly all the patch pairs are normalised so the pixel-and-channel-wise mean and variance are 0 and 1.  We divide the TVS patches $ 80 \% - 20 \% $ to produce, respectively, training and validation sets.  We follow \cite{BIQTCNN,IQTADMRI} in having 8 independent subjects as the test set.  As in \cite{BIQTCNN}, we evaluate our model separately on the interior region RMSE and exterior (peripheral) region RMSE, of the brain.  Furthermore we compare total brain-wise RMSE.    
\\
\\ \noindent \textbf{Implementation} As in \cite{BIQTCNN}, we extract 2250 patches per subject for TVS, where the central voxel lies within its brain-mask.  
We utilise PyTorch \cite{PYTORCH}, He parameter initialisation \cite{DDR}, ADAM optimiser \cite{ADAM} with RMSE loss (between the prediction and the inversely-shuffled HR images), decaying learning rate starting at $ 10^{-4}$, ending training either at 100 epochs or when validation set performance fails to improve.  Given that results vary due to the random initialisation, we train 4 models for each set of hyperparameters, select the best model from the validation score, to then evaluate on the test set.  All our experiments are conducted on a single NVidia Pascal GPU.


\begin{figure}[h]
\centering
\footnotesize
\begin{tabular}{ |c||c|c|c|c| } 
 \hline
    Model                   & Subjects (TVS) & RMSE Interior & RMSE Exterior & RMSE Total \\ 
  \hline \hline
  ESPCN                     & 8             & 6.33 $ ( \pm 0.30 ) $ & 14.01 $ ( \pm 1.12 ) $ & 9.72 $ ( \pm 0.64 ) $    \\ 
  ESPCN-best-\cite{BIQTCNN} & 8             & 6.29 $ ( \pm 0.29 ) $ & 13.82 $ ( \pm 0.31 ) $ & 9.76 $ ( \pm 0.51 ) $   \\
  \hline \hline 
  ESPCN-RN2                 & 8             & 5.78 $ (\pm 0.28) $ & 13.17 $ ( \pm 1.16 ) $ &  9.06 $ ( \pm 0.66 ) $ \\  
  ESPCN-RN4                 & 8             & 5.71 $ (\pm 0.24 ) $ & 12.84 $ (\pm 1.18 ) $ &  8.86 $ (\pm 0.66 ) $   \\  
  ESPCN-RN6                 & 8             & 7.33 $ (\pm 1.43) $ & 13.03 $ (\pm 1.19 ) $ &  9.76 $ (\pm 0.88 ) $   \\ 
  ESPCN-RN8                 & 8             & 9.54 $ (\pm 4.38 ) $ & 12.78 $ (\pm 1.25 ) $ & 11.08 $ (\pm  2.66 ) $   \\  
  \hline \hline
  ESPCN        & 16            & 6.12 $ ( \pm 0.29 ) $ & 13.42 $ ( \pm 1.15 ) $ & 9.33 $ ( \pm 0.65 ) $  \\ 
  ESPCN-RN4    & 16            & 5.51 $ ( \pm 0.25 ) $ & 12.40 $ ( \pm 1.23 ) $ & 8.56 $ ( \pm 0.68 ) $  \\
 \hline
  ESPCN        & 32            & 6.12 $ ( \pm 0.29 ) $ & 13.42 $ ( \pm 1.15 ) $ & 9.33 $ ( \pm 0.65 ) $ \\ 
  ESPCN-RN4    & 32            & 5.58 $ ( \pm 0.25 ) $ & 12.13 $ ( \pm 1.24 ) $ & 8.46 $ ( \pm 0.67 ) $  \\ 
 \hline
 
\end{tabular}
\caption{Comparing mean and std RMSE on 8 test subjects, where we first vary number of RevNet blocks per stack, then size of training-validation set (TVS). Network input size $ 11^{3} $, upsampling rate $r=2 $.}\label{Table_combined}
\end{figure}

\noindent \\ \textbf{IQT Performance}  In Fig.\ref{Table_combined}, increasing network depth improves accuracy, until the models overfit.  Since implementing regularisation deteriorates our results due to the bias-variance tradeoff, we instead utilise larger TVS.  Unlike the ESPCN, our extended model registers improvements on both interior and the exterior (peripheral) brain regions, with additional data.  To assess statistical significance of our results, we employed a non-parametric Wilcoxon signed-rank test (W statistic) for paired RMSE values of our 8 test subjects, comparing our best model (ESPCN-RN4) over state-of-the-art \cite{BIQTCNN} (Fig.\ref{Table_combined}), produces W=0, significant with p=0.0117 (critical value for W is 3 at N=8, at significance level alpha=0.05), improvement of the ESPCN-RN4 over ESPCN at 32 subjects also produces W=0, p=0.0117, which is significant as before.  We note this improvement occurs with almost identical memory usage (Fig.\ref{Table_MemEff}).
\noindent Comparing the image quality enhancement due to our results in Fig.\ref{Table_combined}, we observe from Fig.\ref{Results_pic} that our augmentation produces sharper recovery of anatomical boundaries between white matter and grey matter, whilst better capturing high-frequency details such as textures on white matter. 

\begin{figure}[h]
\begin{center}
\includegraphics[width=\linewidth, height=200pt]{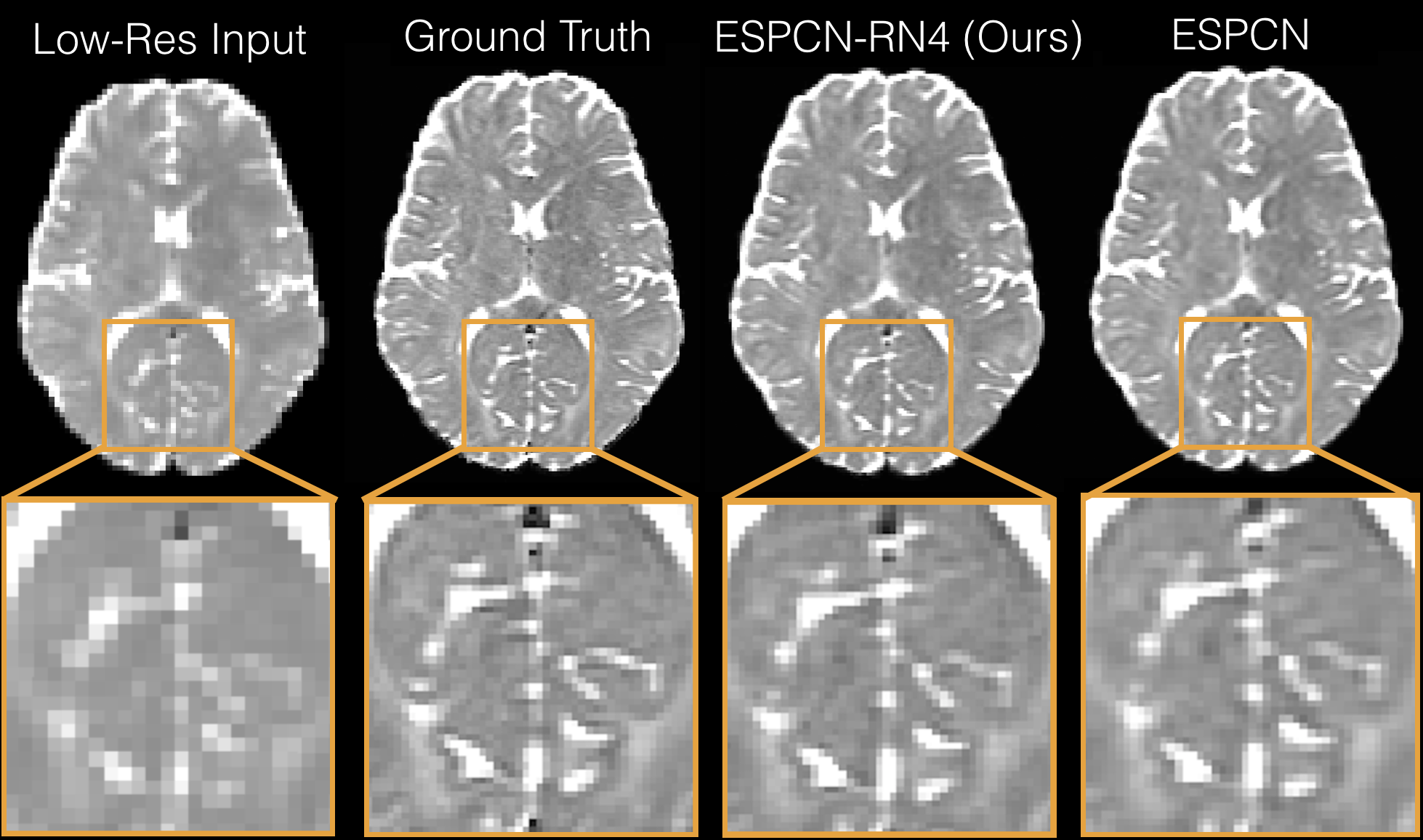}\caption{A visualisation of mean diffusivity maps on an axial slice on a test HCP subject, estimated from: low-resolution input, ground-truth, high-resolution reconstruction from best ESPCN-RN4 (Fig.\ref{Table_combined}), ESPCN \cite{BIQTCNN}.}\label{Results_pic}
\end{center}
\end{figure}
\noindent \\ \textbf{Memory Comparison} Despite significantly elongating the ESPCN, our novel procedure performs very well with respect to memory usage -- an increase of just $ 4.0 \% $ in Fig.\ref{Table_MemEff} -- which also includes caching extra RevNet parameters.  Memory consumption is more than halved when using a low memory scheme (ESPCN-RN4), with respect to naively performing backpropagation from a single computational graph and ignoring both checkpointing and the reversibility property of RevNets (ESPCN-RN4 Naive).  Although the computational time tradeoff of the low-memory system is significant, training for each model in Fig.\ref{Table_combined} was obtained in under 24 hours.  

\begin{figure}[H]
\begin{center}
\footnotesize
\begin{tabular}{ |c||c|c| } 
 \hline
    Model           & Memory Usage (MB) & Computational Time (s)  \\ 
  \hline \hline   
  ESPCN             &  523              &    20       \\ 
  ESPCN-RN4         &  541              &    309       \\ 
  ESPCN-RN4 Naive   & 1091              &    231       \\ 
  \hline
\end{tabular}
\end{center}\caption{Comparing the memory usage and computational time on a single epoch with 8 TVS subjects: the original ESPCN, our augmented ESPCN-RN4 and ESPCN-RN4 Naive (ESPCN-RN4 without the low-memory optimisation). }\label{Table_MemEff}
\end{figure}

\section{Conclusion}

Attempts to merge cutting-edge techniques in deep learning with medical imaging data often encounter memory bottlenecks, due to limited  GPU memory and large data sets.  In this paper, we present how combining checkpointing with RevNets allow us to train long convolutional neural networks with modest computational and memory requirements.  Our example -- dMRI super-resolution in the paradigm of IQT -- illustrates how to improve performance via neural network augmentation, with a negligible increase in memory requirements.  However the benefits of this technique extend to many other applications which use deep neural networks, particularly in medical imaging, where large image volumes are the predominant data type.

\subsection*{Acknowledgements}
This work was supported by an EPRSC scholarship and EPSRC grants M020533 R006032 R014019.  We thank: Adeyemi Akintonde, Tristan Clark, Marco Palombo and Emiliano Rodriguez.  Data were provided by the Human Connectome Project, WU-Minn Consortium (PIs: D. V Essen and K. Ugurbil) funded by NiH and Wash. U.

\bibliographystyle{splncs}
\bibliography{SBB_MICCAI_18}

\end{document}